%% file: main.tex
\documentclass[conference]{IEEEtran}

\IEEEoverridecommandlockouts                              

\usepackage{microtype}
\usepackage{amsmath,amssymb,amsfonts}
\usepackage{algorithmic}
\usepackage{graphicx}
\usepackage{textcomp}
\usepackage{xcolor}
\usepackage{subfig}
\usepackage{gensymb}
\usepackage{caption}
\usepackage{hyperref}
\usepackage{mwe}
\usepackage{amsfonts}
\usepackage{amssymb}
\usepackage{verbatim}
\usepackage{amsbsy}
\usepackage{siunitx}
\usepackage{tabularx, booktabs}
\usepackage{dblfloatfix}
\usepackage{cite}
\usepackage{cleveref}

\usepackage{algorithm, algorithmic}
\usepackage[autolanguage]{numprint}

\usepackage{spreadtab}
\usepackage{multirow}
\usepackage{cite}

\usepackage{hyperref}
\usepackage{xcolor}
\usepackage{algorithmic}
\usepackage{textcomp}
\usepackage{xcolor}
\usepackage{subfig}
\usepackage{gensymb}
\usepackage{caption}

\usepackage{mwe}
\usepackage{amsfonts}
\usepackage{amssymb}
\usepackage{verbatim}
\usepackage{amsbsy}
\usepackage{siunitx}
\usepackage{tabularx, booktabs}
\usepackage{soul}
\usepackage{tikz}
\usetikzlibrary{calc}

\usepackage{graphicx}
\usepackage{subfig} 

\usepackage{geometry}
\makeatletter
\newif\if@anonymize

\@anonymizetrue    

\if@anonymize
  \newcommand{\highlight@DoHighlight}{
    \fill [outer sep = -15pt, inner sep = 0pt, color=black]
          ($(begin highlight)+(0,8pt)$) rectangle ($(end highlight)+(0,-3pt)$) ;
  }

  \newcommand{\highlight@BeginHighlight}{
    \coordinate (begin highlight) at (0,0) ;
  }

  \newcommand{\highlight@EndHighlight}{
    \coordinate (end highlight) at (0,0) ;
  }

  \newdimen\highlight@previous
  \newdimen\highlight@current
  \newlength{\item@width}

  \DeclareRobustCommand*\anonymize{%
    \SOUL@setup
    \def\SOUL@preamble{%
      \begin{tikzpicture}[overlay, remember picture]
        \highlight@BeginHighlight
        \highlight@EndHighlight
      \end{tikzpicture}%
    }%
    \def\SOUL@postamble{%
      \begin{tikzpicture}[overlay, remember picture]
        \highlight@EndHighlight
        \highlight@DoHighlight
      \end{tikzpicture}%
    }%
    \def\SOUL@everyhyphen{%
      \discretionary{%
        \SOUL@setkern\SOUL@hyphkern
        \SOUL@sethyphenchar
        \tikz[overlay, remember picture] \highlight@EndHighlight ;%
      }{%
      }{%
        \SOUL@setkern\SOUL@charkern
      }%
    }%
    \def\SOUL@everyexhyphen##1{%
      \SOUL@setkern\SOUL@hyphkern
      \settowidth{\item@width}{##1}%
      \makebox[\item@width]{}%
      \discretionary{%
        \tikz[overlay, remember picture] \highlight@EndHighlight ;%
      }{%
      }{%
        \SOUL@setkern\SOUL@charkern
      }%
    }%
    \def\SOUL@everysyllable{%
      \begin{tikzpicture}[overlay, remember picture]
        \path let \p0 = (begin highlight), \p1 = (0,0) in \pgfextra
          \global\highlight@previous=\y0
          \global\highlight@current =\y1
        \endpgfextra (0,0) ;
        \ifdim\highlight@current < \highlight@previous
          \highlight@DoHighlight
          \highlight@BeginHighlight
        \fi
      \end{tikzpicture}%
      \settowidth{\item@width}{\the\SOUL@syllable}%
      \makebox[\item@width]{}%
      \tikz[overlay, remember picture] \highlight@EndHighlight ;%
    }%
    \SOUL@
  }
\else
  \newcommand{\anonymize}[1]{#1}
\fi

\newcommand{\linebreakand}{%
  \end{@IEEEauthorhalign}
  \hfill\mbox{}\par
  \mbox{}\hfill\begin{@IEEEauthorhalign}
}

\makeatother 

\title{\LARGE \bf Improving Generalization in Aerial and Terrestrial \\ Mobile Robots Control Through Delayed Policy Learning}

\author{\IEEEauthorblockN{Ricardo B. Grando$^{1}$$^{2}$, Raul Steinmetz$^{3}$, Victor A. Kich$^{4}$, Alisson H. Kolling$^{2}$, Pablo M. Furik$^{1}$,\\Junior C. de Jesus$^{5}$, Bruna V. Guterres$^{1}$, Daniel T. Gamarra$^{3}$, Rodrigo S. Guerra$^{2}$, Paulo L. J. Drews-Jr$^{2}$}
\IEEEauthorblockA{\textit{$^{1}$Universidad Tecnologica del Uruguay}, \textit{$^{2}$Universidade Federal de Rio Grande},\\\textit{$^{3}$Universidade Federal de Santa Maria}, \textit{$^{4}$University  of Tsukuba}, \textit{$^{5}$VersusAI}}
}

\begin{document}

\maketitle

\thispagestyle{empty}
\pagestyle{empty}

\input{sections/0_0_abstract.tex}
\input{sections/0_1_supplementary_material.tex}

\input{sections/1_introduction.tex}
\input{sections/2_related_works.tex}
\input{sections/3_methodology.tex}
\input{sections/4_experimental_results.tex}
\input{sections/5_discussion.tex}
\input{sections/6_conclusion.tex}
\input{sections/7_acknowledgment.tex}
\input{sections/8_references.tex}

\end{document}

%% file: sections/0_0_abstract.tex
\begin{abstract}
Deep Reinforcement Learning (DRL) has emerged as a promising approach to enhancing motion control and decision-making through a wide range of robotic applications. While prior research has demonstrated the efficacy of DRL algorithms in facilitating autonomous mapless navigation for aerial and terrestrial mobile robots, these methods often grapple with poor generalization when faced with unknown tasks and environments. This paper explores the impact of the Delayed Policy Updates (DPU) technique on fostering generalization to new situations, and bolstering the overall performance of agents. Our analysis of DPU in aerial and terrestrial mobile robots reveals that this technique significantly curtails the lack of generalization and accelerates the learning process for agents, enhancing their efficiency across diverse tasks and unknown scenarios.
\end{abstract}


%% file: sections/0_1_supplementary_material.tex


%% file: sections/1_introduction.tex
\section{Introduction}\label{introduction}


Reinforcement Learning (RL) and Deep Reinforcement Learning (DRL) have been used for enhancing autonomous decision-making in machines. RL entails an agent learning to accomplish a goal through actions in an environment, steered by rewards or penalties feedback~\cite{montague1999reinforcement}. DRL, integrating deep neural networks with RL, advances these capabilities, enabling the approximation of optimal policies for handling complex inputs and learning from vast amounts of unstructured data. These advancements have spurred significant progress in areas such as gaming~\cite{mnih2015human} and robotics~\cite{gu2016deep, gu2017deep}. However, a notable challenge in model-free DRL is agents' susceptibility to poor generalization in unfamiliar scenarios, limiting their task generalization effectiveness~\cite{packer2018assessing}.


Generalization, the capacity of neural networks to apply learned knowledge to unseen data, is essential for neural network development, representing a primary objective in deep learning~\cite{novak2018sensitivity}. Insufficient generalization limits a model's usefulness, resulting in subpar performance on data not encountered during training. Various techniques have been proposed to bolster generalization, such as dropout~\cite{srivastava2014dropout, wager2013dropout}, normalization~\cite{ioffe2015batch, klambauer2017self}, and ensemble methods~\cite{freund1997decision}, enhancing model robustness and effectiveness.

\begin{figure}[htbp]
    \centering \includegraphics[width=\linewidth]{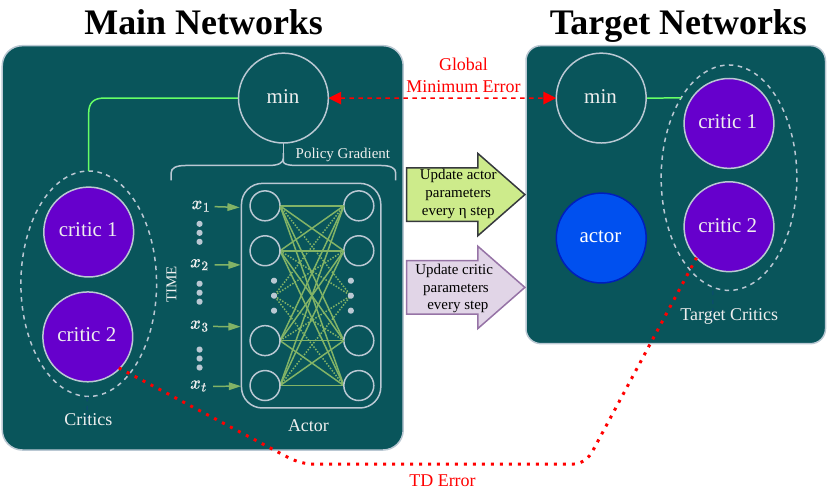}
    \caption{Architecture of the Twin Delayed Deep Deterministic Policy Gradient (TD3) algorithm and its mechanisms.}
    \label{fig:td3-dpu-diagram}
    \vspace{-5mm}
\end{figure}

Improving generalization in DRL is critical for ensuring robust performance across a wide range of unknown scenarios. This is particularly important as DRL agents often excel in their training environments, optimizing their behavior to such an extent that they appear to overfit these specific scenarios. However, when faced with changes in the environment, as small as they may be (a new object to avoid in autonomous navigation for example), their performance can drastically decline, leading to a significant drop in task accuracy and their ability to complete it successfully~\cite{zhang2018study}. Numerous methods have been developed to enhance generalization in deep reinforcement learning, including soft data augmentation~\cite{hansen2021generalization}, mixture regularization~\cite{wang2020improving}, and others. 

 
Delayed Policy Updates (DPU)~\cite{fujimoto2018addressing} is one of these techniques and it aims to bolster learning stability by postponing the agent's policy updates, allowing for learning from a steadier experience pool over extended periods. This study focuses on analyzing DPU's effect on RL agents' generalization in navigation tasks for mobile robots, a critical area where robots must adapt to new tasks or environments while retaining learned knowledge. We demonstrate through experiments with varying DPU settings in different environments that DPU significantly enhances generalization. Our research evaluates DPU's role in addressing generalization challenges within RL for both aerial and terrestrial navigation of robots.

The present work specifically contributes on the following topics:


\begin{itemize}
\item We train three identical agents to navigate autonomously as aerial robots, and three to navigate autonomously as terrestrial robots, each configured with different DPU values. Our analysis reveals that a greater delay contributes positively to the success rate for a navigation task assigned to an aerial or mobile robot.
\item In a subsequent experiment within an unfamiliar scenario, we confirm the agents' training effectiveness. The results demonstrate that an increased delay enhances greatly the agent's generalization capabilities across different environments in both applications.
\end{itemize}

The work has the following structure: related works are addressed in the Section~\ref{related_works}. Our approaches and tools used can be seen in the Section~\ref{methodology}. The results achieved are presented in the Section~\ref{results}. Section~\ref{discussion} shows our overall analysis of the results and Section~\ref{conclusions} presents the perspective for future works.

%% file: sections/2_related_works.tex
\section{Related Work}\label{related_works}

Throughout the years, a vast number of techniques have been developed to enhance the generalization capabilities of DRL architectures. Karl Cobbe \emph{et al.}~\cite{cobbe2019quantifying} delved into the issue of overfitting within deep reinforcement learning, emphasizing the difficulty of achieving generalization across procedurally generated environments. By introducing CoinRun, a benchmark aimed at evaluating RL generalization, they revealed that agents often exhibit overfitting when trained on extensive datasets. Their findings suggested that adopting deeper convolutional structures and incorporating methods commonly used in supervised learning, such as $\ell2$ regularization, dropout, data augmentation, and batch normalization, can significantly bolster generalization in RL contexts.

Building on this understanding, Farebrother \emph{et al.}~\cite{farebrother2018generalization} also tackled the challenge of overfitting in DRL, focusing on DQN models applied to Atari 2600 games. They underscored the innovative application of game modes as a means to assess the generalization capacity of DQN agents, showcasing how the integration of regularization techniques like dropout and $\ell2$ regularization can amplify the adaptability of DQN models across varied game scenarios. This work provides an additional layer of insight into addressing the generalization dilemma within RL environments.

In a complementary effort, Lee \emph{et al.}~\cite{lee2019network} proposed a novel method aimed at refining the generalization of DRL agents. Their technique, which involves the utilization of randomized (convolutional) neural networks to alter input observations, encourages the formation of robust features that remain consistent across a spectrum of randomized environments. Moreover, they introduced a Monte Carlo approximation-based inference strategy to reduce the variance associated with this input perturbation, further contributing to the toolkit for enhancing RL generalization.

Extending the exploration of generalization in DRL, Justesen \emph{et al.}~\cite{justesen2018illuminating} investigated the efficacy of using procedurally generated levels during training. Their research demonstrated that such an approach not only facilitates the generalization of RL models to new, similarly distributed levels but also improves efficiency by dynamically adjusting the difficulty of levels in response to the agent's performance, highlighting a strategic method to optimize training processes.

Further advancing the field, Raileanu \emph{et al.}~\cite{raileanu2020automatic} introduced automated techniques for identifying effective data augmentations across various tasks, markedly improving agent generalization. This innovation circumvents the need for expert knowledge in choosing task-specific augmentations, addressing a significant challenge in scaling RL applications. Additionally, the introduction of two novel regularization terms for policy and value functions integrates data augmentation into actor-critic algorithms at a theoretical level, marking a significant step forward in the quest for enhanced RL generalization.

The Twin Delayed Deep Deterministic Policy Gradient (TD3) algorithm, introduced by Fujimoto \emph{et al.}~\cite{fujimoto2018addressing}, advances the Deep Deterministic Policy Gradient (DDPG) framework~\cite{lillicrap2015continuous} for improved performance in continuous control tasks within RL. By employing dual critics (critic 1 and critic 2) for Q-value estimation, TD3 strategically addresses the overestimation bias common in Q-learning algorithms. A key innovation of TD3 is the asynchronous update of policy (actor) and Q-value estimators (critics), introducing a deliberate delay in policy updates to mitigate the risk of overfitting to premature Q-value estimates. This methodological refinement has seen application across various domains, notably in enhancing mobile robotics tasks~\cite{gao2020deep, grando2022double, grando2022mapless, grando2023docrl, li2022path}, these works focus primarily on algorithmic or task-specific improvements rather than on the learning impact of the DPU technique itself.

This study aims to fill a gap in the literature by analyzing the effects of DPU within the context of continuous control in RL, particularly concerning learning generalization. We delve into how delayed policy updates influence learning processes and an agent's capability to execute specific tasks, with a focus on navigation tasks for aerial an terrestrial mobile robots. Which to the best of our knowledge, consists in a novel application not yet explored in existing research.

%% file: sections/3_methodology.tex
\section{Methodology}\label{methodology}

In this section, we present our DRL approach. We detail the network structure and the implementation of the TD3 algorithm. We also present the simulation details, and the task used to perform the analysis.




\subsection{Twin Delayed Deep Deterministic Policy Gradient}

The Deep Deterministic Policy Gradient (DDPG) algorithm, introduced by Lillicrap \emph{et al.}~\cite{lillicrap2015continuous}, marked a significant advancement in the field of reinforcement learning for continuous control tasks, particularly in robotics. DDPG, inspired by the success of Q-learning and its adaptations for continuous action spaces, integrates an actor-critic architecture to balance the policy performance (actor) with the value function approximation (critic). The actor network generates continuous actions, while the critic network provides a stable target value, enhancing the algorithm's stability and making it well-suited for applications in mobile robotics~\cite{jesus2019deep}. Despite its achievements, DDPG is prone to overestimating Q-values, which can destabilize the learning policy and lead to suboptimal performance.

\begin{figure}[tbp]
    \centering
    \subfloat[First scenario.]{\includegraphics[width=0.48\linewidth]{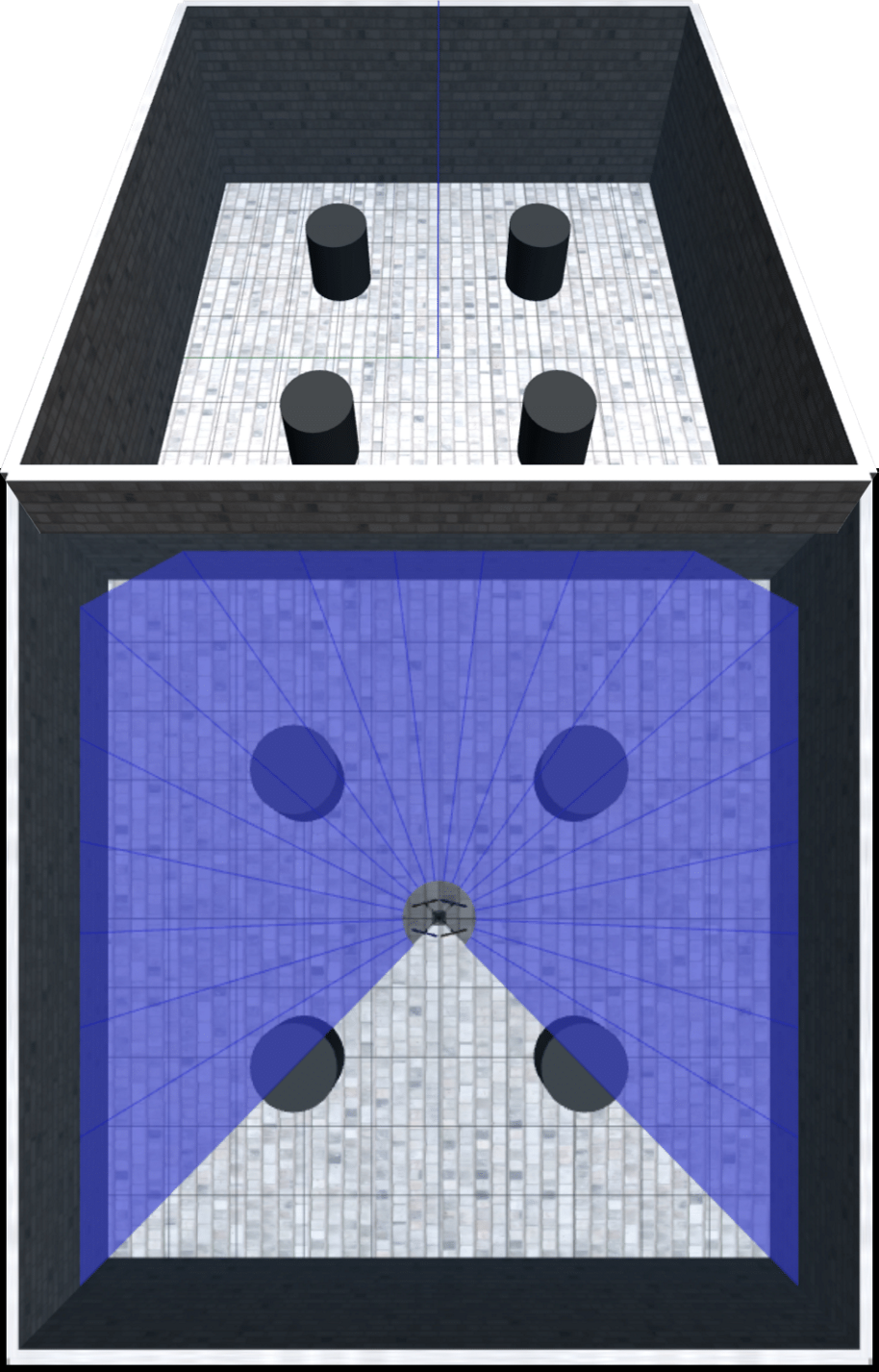}\label{fig:stage1}}
    \hspace{1mm} 
    \subfloat[Second scenario.]{\includegraphics[width=0.48\linewidth]{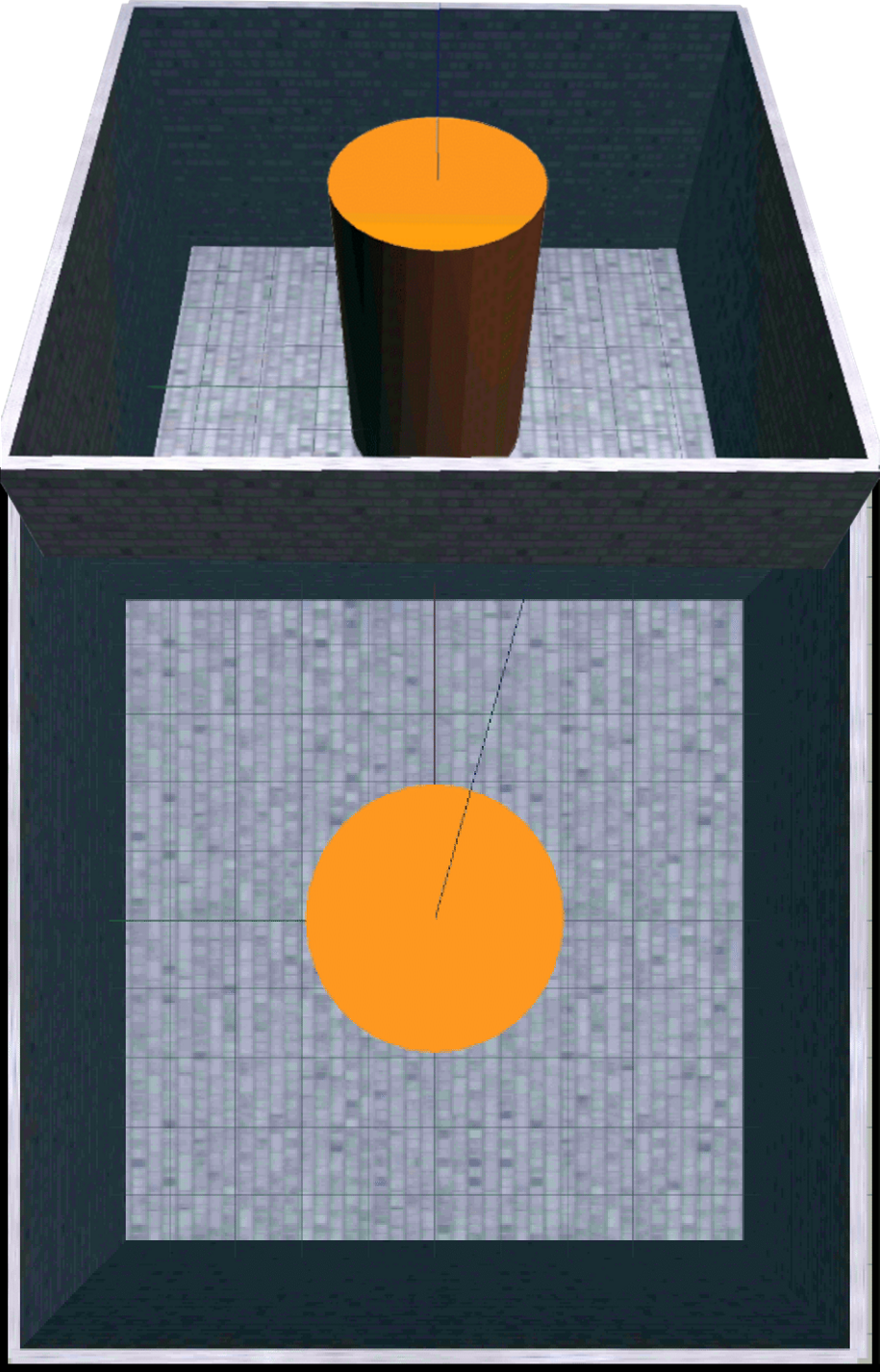}\label{fig:stage2}}
    \caption{Training and evaluation scenarios for aerial mobile robots, respectively.}
    \label{fig:stages}
    \vspace{-5mm}
\end{figure}

To address these challenges, the TD3 algorithm, introduces several key improvements over its predecessor. By employing clipped double-Q learning, TD3 mitigates the overestimation bias by using two critic networks to calculate the Bellman error loss functions, taking the minimum value of their Q-value estimations. Furthermore, it incorporates target policy smoothing by adding Gaussian noise to the target action, enhancing exploration and preventing policy exploitation. A distinctive feature of TD3 is its delayed policy updates strategy, wherein the policy network's updates are less frequent than those of the critic networks. The delay parameter $\eta$ represents the number of environment steps that should be executed before updating the target actor network for the learning calculations. This delay reduces the risk of policy overfitting, thereby promoting more stable learning and better generalization across a wider range of tasks, as shown in Section~\ref{results}. The implementation of the TD3 algorithm in this work incorporates these advancements, demonstrating its efficacy in complex control scenarios. The Algorithm~\ref{alg:docrl_d} presents the TD3 algorithm implemented for this work, and the diagram of this architecture is depicted in Fig.\ref{fig:td3-dpu-diagram}.


\begin{algorithm}[!bp]
    \algsetup{linenosize=\tiny}
    \scriptsize
    \caption{TD3 Algorithm}
    \label{alg:docrl_d}
    \begin{algorithmic}[1]
        \STATE Initialize params of critic networks $\theta_{1}$, $\theta_{2}$ , and actor-network $\pi(\phi)$
        \STATE Initialize params of target networks $\phi^{\prime}\leftarrow\phi$, $\theta_{1}^{\prime}\leftarrow\theta_{1}$, $\theta_{2}^{\prime}\leftarrow\theta_{2}$
        \STATE Initialize replay buffer $\beta$
        \FOR{$global\_step$ to $max\_global\_steps$}
            \WHILE{$episode$ not $done$}
                \STATE reset environment state
                \FOR{$t = 0$ to $max\_steps$}
                    \IF {$t < start\_steps$}
                        \STATE $a_{t} \leftarrow $ env.action\_space.sample() 
                    \ELSE
                        \STATE $a_{t}\leftarrow\pi_{\phi}(s)+\epsilon,\ \epsilon\sim \mathcal{N}(0,OU)$
                    \ENDIF
                    
                    \STATE $s_{t+1}$, $r_{t}$, $d_{t}$, \_ $\leftarrow$ env.step($a_{t}$)
                    
                    \STATE store the new transition $(s_{t}, a_{t}, r_{t}, s_{t+1}, d_{t})$ into $\beta$
                    
                    \IF{$t > start\_steps$}
                        \STATE Sample mini-batch of $N$ transitions $(s_{t}, a_{t}, r_{t},s_{t+1}, d_{t})$ from $\beta$
                        
                        \STATE $a'\leftarrow\pi_{\phi^{\prime}}(s^{\prime})+\epsilon,\ \epsilon\sim clip(\mathcal{N}(0,\tilde{\sigma}), -c,\ c)$ 
                        
                        \STATE Computes target: \\ $Q_{t} \leftarrow r+\gamma*\min_{i=1,2}Q_{\theta_i}(s', a')$
                        
                        
                        \STATE Update double critics with one-step gradient descent:\\
                       $\nabla_{\theta_i} \frac{1}{N} \sum_i(Q_t - Q_{\theta_{i}(s_{t},a_{t})})^2$  \qquad for i=1,2
                        
                        \IF {t \% $\eta$ == 0}
                            \STATE Update policy with one-step gradient descent:\\             
                            $\nabla_{\phi}\frac{1}{N} \sum_i[\nabla_{a_{t}}Q_{\theta_{1}}(s_{t},a_{t})\vert _{a_{t}=\pi(\phi)}\nabla_{\phi}\pi_{\phi}(s_{t})]$
                            Soft update for the target networks: \\
                            \STATE $\phi^{\prime}\leftarrow\tau\phi+(1-\tau)\phi^{\prime}$
                            \STATE $\theta_{i}^{\prime}\leftarrow\tau\theta_{i}+(1-\tau)\theta_{i}^{\prime}$ \qquad for i=1,2
                        \ENDIF
                    \ENDIF
                \ENDFOR
            \ENDWHILE
        \ENDFOR
  \end{algorithmic}
\end{algorithm}

\begin{figure}[tbp]
    \centering
    \subfloat[First scenario.]
    {\includegraphics[width=0.48\linewidth]{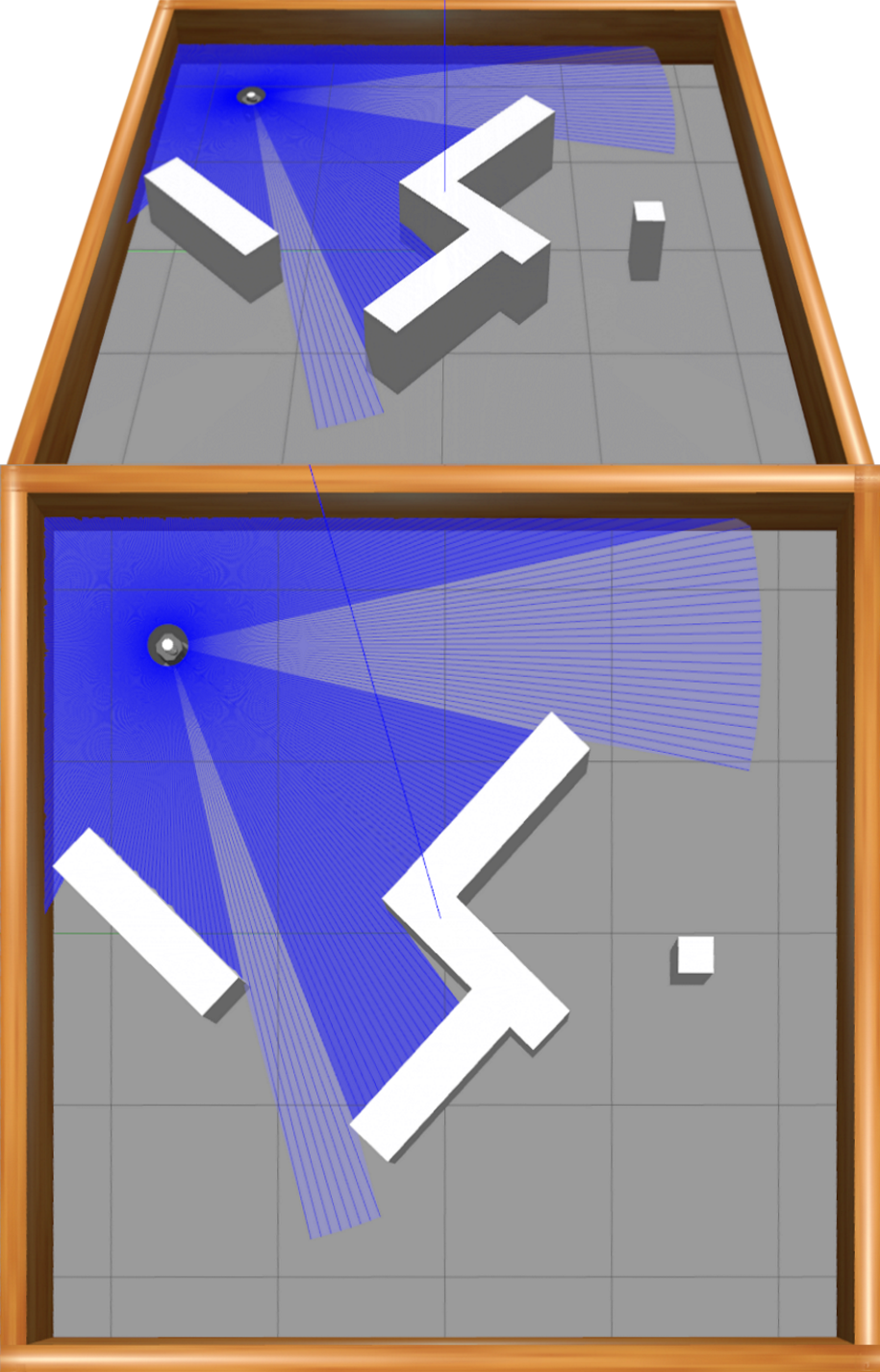}\label{fig:stage3}}
    \hspace{1mm} 
    \subfloat[Second scenario.]
    {\includegraphics[width=0.48\linewidth]{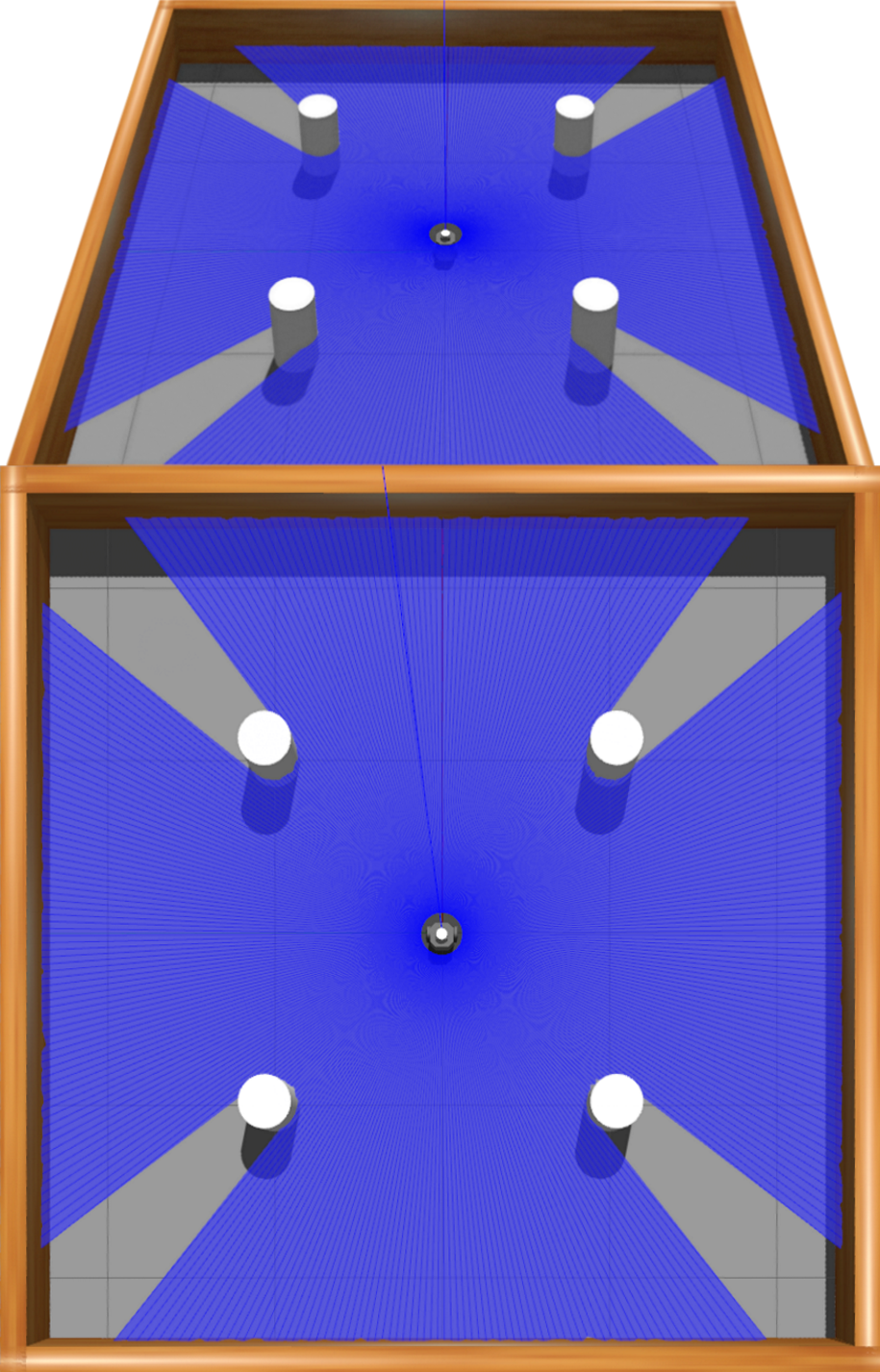}\label{fig:stage4}}
    \hspace{1mm} 
    \caption{Training and evaluation scenarios for terrestrial mobile robots, respectively.}
    \label{fig:terrestrial_stages}
    \vspace{-5mm}
\end{figure}

We use a fully connected ANN as actor-network and denote them by $\phi$ and its copy $\phi^{\prime}$ as actor target. The actor target chooses the action $a^{\prime}$ based on the state $s^{\prime}$, and we add Ornstein-Uhlenbeck~\cite{nauta2019using} noise to it. The double critic targets take the tuple ($s^{\prime}$, $a^{\prime}$) and return two Q-values as output. The minimum of the two target Q-values is considered as the approximated value return. The loss is calculated with the Mean Squared Error of the approximate value from the target networks and the value from the critic networks. We use Adaptive Moment Estimation (Adam) to minimize the loss.

Three variants of the algorithm were used in the methodology. The first one sets the parameter $\eta=2$, the second one uses $\eta=4$ and the last one uses $\eta=8$. Each agent was trained and evaluated independently.

\subsection{Aerial Robot Simulated Environments and Task Description}

Our research utilized aerial mobile robots in simulations developed by Grando \emph{et al.}~\cite{grando2022double} and Jesus \emph{et al.}~\cite{de2022depth}, utilizing the Gazebo simulator in conjunction with ROS. The RotorS framework~\cite{furrer2016rotors} served as the foundation, facilitating the simulation of aerial vehicles with capabilities for various command levels including angular rates, attitude, and location control, as well as wind simulation using the Ornstein-Uhlenbeck process.

\begin{figure}[!tp]
    \centering
    \includegraphics[width=\linewidth]{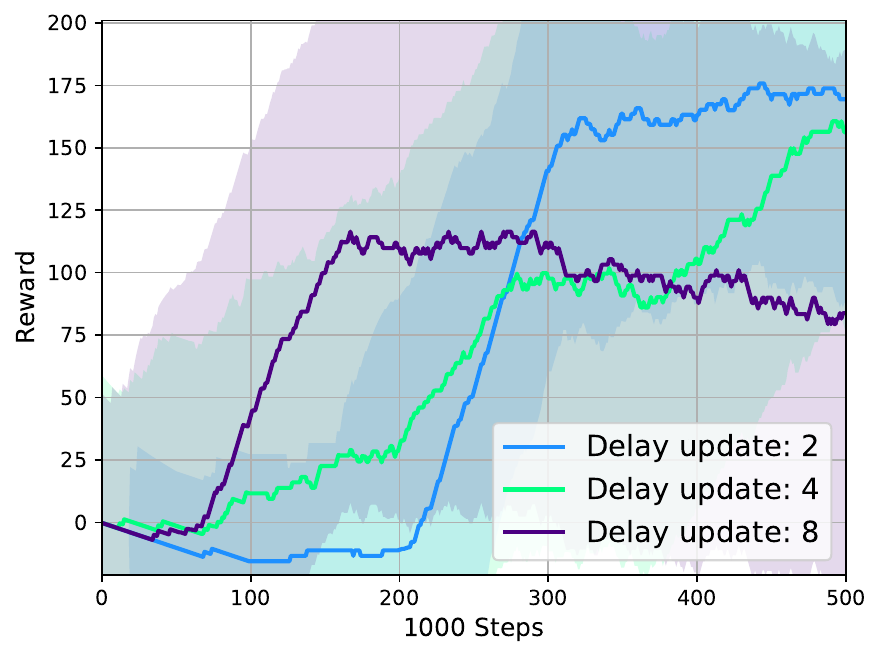}
    \caption{Reward moving average of the aerial mobile robot over 500 episodes of training.}
    \label{fig:rewards}
    \vspace{-5mm}
\end{figure}

We utilized two 10-meter square scenarios: the first features four symmetrically placed obstacles around the center, and the second contains a single central obstacle. The first scenario was used for training purposes, while the second evaluated the model's generalization capabilities and retention of learned behaviors. Fig.~\ref{fig:stages} depicts both scenarios.

In each scenario, the robot was tasked with navigating from a start point to a goal, requiring obstacle detection and avoidance, facilitated by LIDAR data. The environment state provided to the agent encompasses 26 dimensions: 20 from distance sensors, three from previous actions, and three detailing the target's position (including relative position and angles in the x-y and z-range planes). Distance data was collected via LIDAR, spaced at 13.5-degree intervals across a 270-degree field.

Action scales range from 0 to 0.25 m/s for linear velocity, -0.25 to 0.25 m/s for altitude adjustments, and -0.25 to 0.25 radians for angular changes ($\Delta$ yaw), enabling effective navigation and obstacle avoidance.

A binary reward function was employed, offering a positive reward for successful navigation and a negative reward for collisions or reaching the 500-step episode limit:

\begin{equation}
r(s_t, a_t)= 
\begin{cases}
    r_{arrive}           & \text{if } d_t < c_d\\
    r_{collide}          & \text{if } min_x < c_o\ ||\ ep = 500,\\
\end{cases}
\end{equation}
\noindent where $r_{arrive}=200$ is given to the agent when completed the task successfully, while $r_{collide}=-20$ penalizes collisions or timeouts, with both $c_d$ and $c_o$ set to 0.5 meters.

\subsection{Terrestrial Robot Simulated Environments and Task Description}

Our experiments were conducted using a terrestrial mobile robot within simulated environments, utilizing the Gazebo simulator in conjunction with ROS2 and the Turtlebot3 robot.

We utilized two 5-meter square scenarios: the first presented three obstacles that constrained the robot's movement through the center, while the second scenario arranged four obstacles symmetrically around the center. The first scenario was used for training purposes, while the second evaluated the model's generalization capabilities and retention of learned behaviors. Fig.~\ref{fig:terrestrial_stages} depicts both scenarios.

Like the aerial simulations, the terrestrial robot's task involved navigating from a start to a goal, relying on LIDAR for obstacle detection. The environment state fed to the agent included 14 dimensions: Ten sensor distance readings, distance and angle to target, and the robot's current linear and angular velocity. Action scales range from 0 to 0.25 m/s for linear velocity and -0.25 to 0.25 m/s for angular velocity.

A binary reward function was employed, offering a positive reward for successful navigation and a negative reward for collisions or reaching the 250-step episode limit:

\begin{equation}
r(s_t, a_t)= 
\begin{cases}
    r_{arrive}           & \text{if } d_t < c_d\\
    r_{collide}          & \text{if } min_x < c_o\ ||\ ep = 250,\\
\end{cases}
\end{equation}
\noindent where $r_{arrive}=100$ is given to the agent when completed the task successfully, while $r_{collide}=-10$ penalizes collisions or timeouts, with $c_d$ set to 0.3 meters and $c_o$ set to 0.19 meters.

%% file: sections/4_experimental_results.tex
\section{Experimental Results}\label{results} 

Regarding the aerial task, the reward trend over 500 episodes, depicted in Fig.~\ref{fig:rewards}, illustrates that agents with higher DPU values not only learn more rapidly but also attain a stabilized reward pattern sooner. This early stabilization suggests that higher DPU values enable agents to efficiently balance exploration with exploitation, although they may not reach the maximal reward potential obtained by agents with lower DPU values that exhibit more specialized behavior in the known scenario.

\begin{figure}
    \centering
    \includegraphics[width=\linewidth]{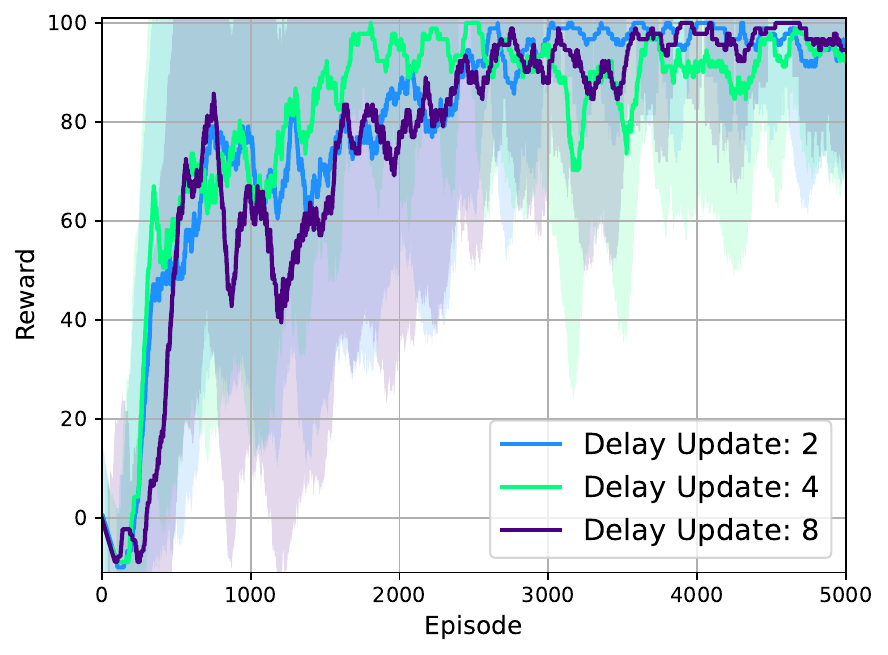}
    \caption{Reward moving average of the terrestrial mobile robot over 5000 episodes of training.}
    \label{fig:training-terrestrial}
    \vspace{-5mm}
\end{figure}

In the context of terrestrial robot navigation, the observed reward trend exhibited a high degree of similarity across all values of DPU, with each agent successfully attaining the maximum reward and sustaining knowledge retention, as depicted in Fig.~\ref{fig:training-terrestrial}. Remarkably, towards the conclusion of the experiments, it was observed that the systems characterized by $\eta = 2$ and $\eta = 4$ exhibited significant fluctuations in reward outcomes, demonstrating frequent failures. This phenomenon is further illustrated by the moving average, calculated over a reduced interval, which reveals the green (representing $\eta=4$) and blue (representing $\eta=2$) lines exhibiting pronounced oscillations in performance towards the latter stages of the study, in stark contrast to the purple line (representing $\eta=8$), which maintains a consistent level of stability.

 To test agent performance under various DPU conditions, we employ metrics such as Success Rate, Episode Reward (ER) Mean and Standard Deviation (Std.), along with Episode Time (ET) Mean and Std. The compiled outcomes for scenarios both recognized and unfamiliar are presented in Tables~\ref{tab:network-metrics-known-aerial} and~\ref{tab:network-metrics-unknown-aerial} for aerial tasks, as well as in Tables~\ref{tab:network-metrics-known-terrestrial} and~\ref{tab:network-metrics-unknown-terrestrial} for terrestrial tasks. Analysis of the data reveals that elevated DPU values enhance performance in known environments while significantly boosting performance in unknown settings (suggesting that lower DPUs may tend to overfit). This is exemplified by the observation that with $\eta=2$, performance in unfamiliar environments decreases dramatically, from 82.98\% to 16.33\% in aerial tasks, and from 93.94\% to 70.71\% in terrestrial tasks. Conversely, a DPU of $\eta=8$ demonstrates a minor decline from 100\% to 98.67\% in aerial scenarios and from 99\% to 85\% in terrestrial scenarios, thereby illustrating superior generalization capabilities. 

 \begin{table}[tp]
    \centering
    \setlength{\tabcolsep}{8.3pt}
    \caption{Aerial mobile robot metrics for different updates delay ($\eta$) in known scenarios.}
    \begin{tabular}{cccccc}
        \toprule
        $\eta$ & \begin{tabular}[c]{@{}c@{}}Success\\Rate (\%)\end{tabular} & \begin{tabular}[c]{@{}c@{}}ER Mean\end{tabular} & \begin{tabular}[c]{@{}c@{}}ER Std.\end{tabular} & \begin{tabular}[c]{@{}c@{}}ET Mean\end{tabular} & \begin{tabular}[c]{@{}c@{}}ET Std.\end{tabular} \\
        \midrule
        2 & 82.98\% & 162.55& 83.12 & 33.53 & 12.86 \\
        4 & 100.00\% & 200.00 & 0.00 & 27.25 & 5.90 \\
        8 & 100.00\% & 200.00 & 0.00 & 22.89 & 1.64 \\
        \bottomrule
    \end{tabular}
    \label{tab:network-metrics-known-aerial}
    \vspace{-5mm}
\end{table}

\begin{table}[bp]
    \vspace{-5mm}
    \centering
    \setlength{\tabcolsep}{8.4pt}
    \caption{Aerial mobile robot metrics for different updates delay ($\eta$) in unknown scenarios.}
    \begin{tabular}{cccccc}
        \toprule
        $\eta$ & \begin{tabular}[c]{@{}c@{}}Success\\Rate (\%)\end{tabular} & \begin{tabular}[c]{@{}c@{}}ER Mean\end{tabular} & \begin{tabular}[c]{@{}c@{}}ER Std.\end{tabular} & \begin{tabular}[c]{@{}c@{}}ET Mean\end{tabular} & \begin{tabular}[c]{@{}c@{}}ET Std.\end{tabular} \\
        \midrule
        2 & 16.33\% & 21.63 & 79.67 & 45.43 & 40.05 \\
        4 & 73.68\% & 142.10 & 97.39 & 22.33 & 10.12 \\
        8 & 98.67\% & 197.07 & 25.40 & 34.60 & 21.65 \\
        \bottomrule
    \end{tabular}
    \label{tab:network-metrics-unknown-aerial}
\end{table}

\begin{table}[tp]
\centering
\setlength{\tabcolsep}{19.5pt}
\caption{Terrestrial mobile robot metrics for different updates delay ($\eta$) in known scenarios.}
\begin{tabular}{cccc}
\toprule
$\eta$ & \begin{tabular}[c]{@{}c@{}}Success\\Rate (\%)\end{tabular} & \begin{tabular}[c]{@{}c@{}}ER Mean\end{tabular} & \begin{tabular}[c]{@{}c@{}}ER Std.\end{tabular} \\
\midrule
2 & 94 & 93.4 & 26.12 \\
4 & 97 & 96.7 & 18.76 \\
8 & 99 & 98.9 & 10.94 \\
\bottomrule
\end{tabular}
\label{tab:network-metrics-known-terrestrial}
\vspace{-5mm}
\end{table}

\begin{table}[bp]
\vspace{-5mm}
\centering
\setlength{\tabcolsep}{19.5pt}
\caption{Terrestrial mobile robot metrics for different updates delay ($\eta$) in unknown scenarios.}
\begin{tabular}{cccc}
\toprule
$\eta$ & \begin{tabular}[c]{@{}c@{}}Success\\Rate (\%)\end{tabular} & \begin{tabular}[c]{@{}c@{}}ER Mean\end{tabular} & \begin{tabular}[c]{@{}c@{}}ER Std.\end{tabular} \\
\midrule
2 & 71 & 68.1 & 49.91 \\
4 & 72 & 69.2 & 49.39 \\
8 & 85 & 83.5 & 39.28 \\
\bottomrule
\end{tabular}
\label{tab:network-metrics-unknown-terrestrial}
\end{table}

\begin{figure*}[!ht]
    \centering
    \subfloat[Trajectory for $\eta=2$.]{\includegraphics[width=0.33\linewidth]{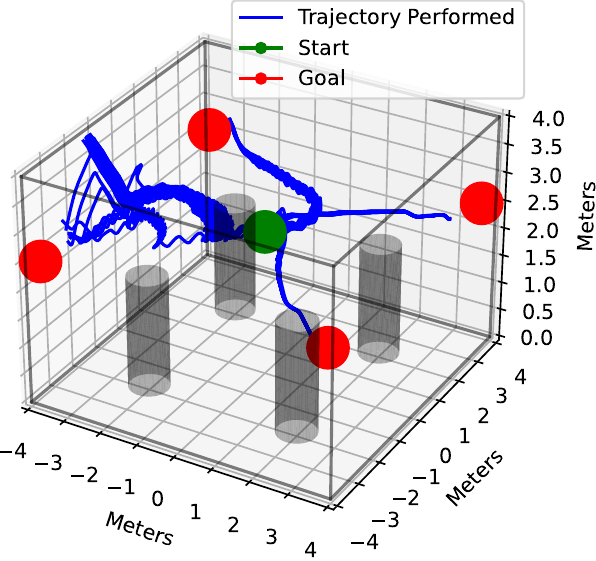}\label{fig:sub11}}
    \subfloat[Trajectory for $\eta=4$.]{\includegraphics[width=0.33\linewidth]{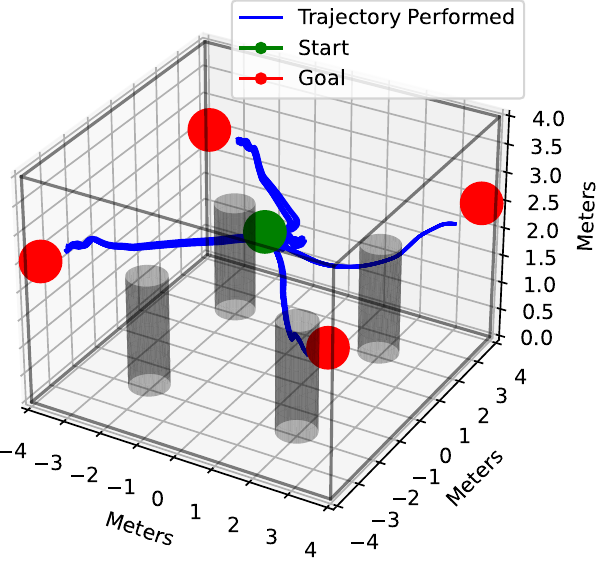}\label{fig:sub12}}
    \subfloat[Trajectory for $\eta=8$.]{\includegraphics[width=0.33\linewidth]{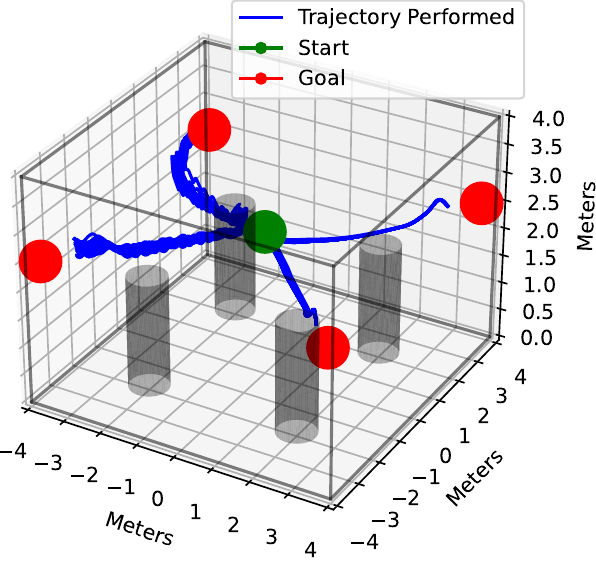}\label{fig:sub13}}
    \caption{Trajectories achieved by the drone controlled by TD3 with different update delays ($\eta$) in known scenario.}
    \label{fig:environment1}
    \vspace{-5mm}
\end{figure*}

\begin{figure*}[!bp]
    \vspace{-5mm}
    \centering
    \subfloat[Trajectory for $\eta=2$.]{\includegraphics[width=0.33\textwidth]{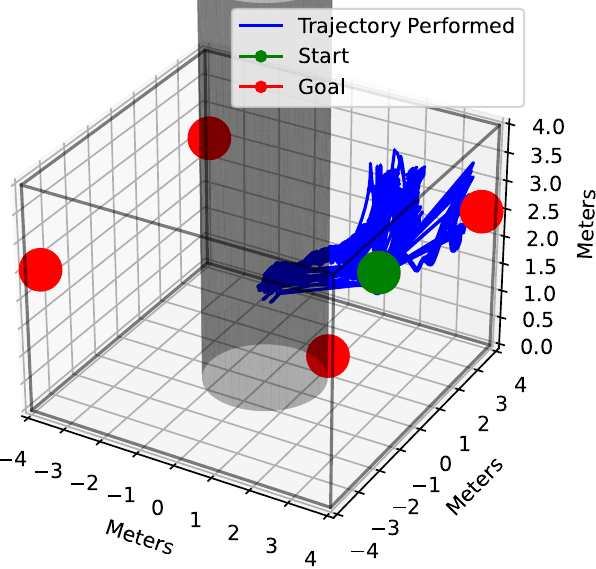}\label{fig:sub21}}
    \subfloat[Trajectory for $\eta=4$.]{\includegraphics[width=0.33\textwidth]{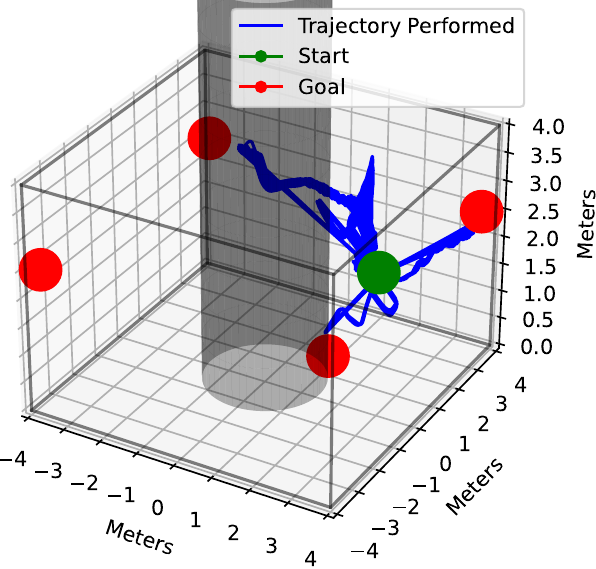}\label{fig:sub22}}
    \subfloat[Trajectory for $\eta=8$.]{\includegraphics[width=0.33\textwidth]{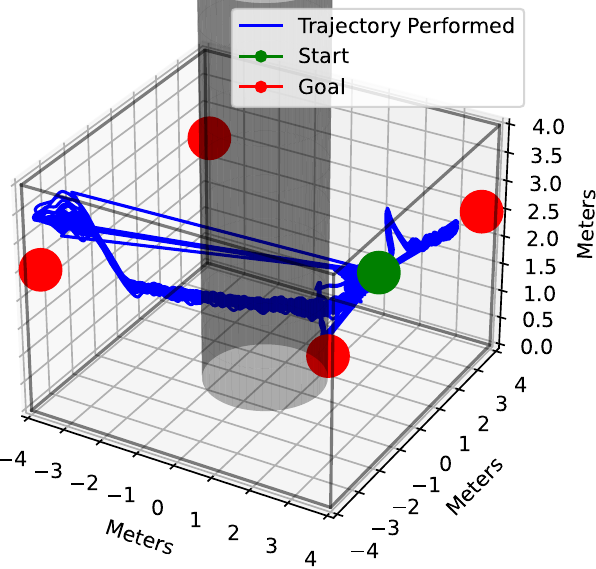}\label{fig:sub23}}
    \caption{Trajectories achieved by the drone controlled by TD3 with different update delays ($\eta$) in unknown scenario.}
    \label{fig:environment2}
\end{figure*}

Analysis of the aerial agent trajectories in the known environment (Fig.~\ref{fig:environment1}) and the unknown environment (Fig.~\ref{fig:environment2}) provides valuable insights into the agents' navigation strategies. In the known scenario, trajectories for higher DPU values ($\eta=4$ and $\eta=8$) reveal more direct and efficient paths to the goal, consistent with their 100\% success rates. In contrast, the agent with a lower DPU value ($\eta=2$) exhibits a more exploratory and less optimized path, correlating with a lower success rate of 82.98\%.

When examining trajectories in the unknown environment, the higher DPU agent ($\eta=8$) demonstrates remarkable adaptability, maintaining a clear and concise path to the goal, as illustrated in Fig.~\ref{fig:sub23}, which is in line with its high success rate of 98.67\%. Conversely, the agent with a lower DPU value ($\eta=2$) shows significant difficulty in navigating the new environment, with erratic movements and a marked decrease in success rate to 16.33\%, further emphasizing the importance of DPU in unfamiliar settings, and the fact that low delays cause dramatic lack of generalization.

%% file: sections/5_discussion.tex
\section{Discussion}\label{discussion}

The experimental findings illuminate the significant impact of Delayed Policy Updates on the learning dynamics and adaptability of reinforcement learning agents. Specifically, a higher DPU value correlates with both accelerated learning in initial training phases and enhanced generalization to new scenarios. This suggests that delaying policy updates enables agents to develop more robust and flexible strategies that are effective across a variety of environmental conditions.

The superior generalization observed with higher DPU values, particularly the minimal performance degradation when transitioning from known to unknown scenarios, underscores the value of this approach in applications where agents must operate in diverse and dynamically changing environments, as noted in Tables \ref{tab:network-metrics-unknown-aerial} and ~\ref{tab:network-metrics-unknown-terrestrial}. It contrasts with the tendency of agents trained with lower DPU values to become highly specialized to their training environments, which can limit their effectiveness in new or altered contexts.

This study’s insights have important implications for designing and training reinforcement learning agents, especially in domains requiring a balance between high performance in familiar settings and the ability to adapt to new situations. Optimizing the DPU parameter emerges as a crucial strategy for achieving this balance, providing a mechanism to tailor agent behavior to meet the specific challenges of the task at hand.

In conclusion, our work contributes to a deeper understanding of how Delayed Policy Updates influence the development of more adaptable and generalizable reinforcement learning agents, paving the way for future research to explore the optimal configuration of these parameters in various application domains.

%% file: sections/6_conclusion.tex
\section{Conclusions}\label{conclusions}

This study has illuminated the significant role that DPU play in the domain of reinforcement learning, particularly within the ambit of robotic navigation. Our empirical analysis has underscored that DPUs exert a pronounced influence on not just the learning speed of RL agents, but also their ability to generalize across diverse environments and optimize their navigation trajectories. A key takeaway from our experiments is the demonstrable advantage of higher DPU values in contexts that demand rapid adaptation from the agents.

Our investigation into the Delayed Policy Updates technique has shed light on its potential to enhance the continuous control capabilities of RL agents engaged in complex navigation tasks. Notably, we found that an agent configured with a bigger DPU value ($\eta=8$) exhibited a negligible drop in success rate when transitioning to an unfamiliar environment, a stark contrast to the performance of the agent with a lower DPU value ($\eta=2$), which saw a dramatic decrease. These insights are pivotal, as they reveal how subtle adjustments in the learning algorithm's parameters can yield substantial differences in an agent's adaptability and competency.

In looking forward, we are poised to delve deeper into the nuances of DPU effects, intending to explore a broader spectrum of DPU values. Furthermore, our future endeavors will aim to dissect and understand other techniques that may interact with DPUs, particularly in addressing task generalization in reinforcement learning. By expanding our scope, we anticipate uncovering richer insights that could lead to more robust and versatile learning algorithms, tailored to the multifaceted demands of real-world applications.

%% file: sections/7_acknowledgment.tex
\section*{ACKNOWLEDGEMENT}

This work was partly funded by the Coordenação de Aperfeiçoamento de Pessoal de Nível Superior (CAPES, brazil), Conselho Nacional de Desenvolvimento Científico e Tecnológico (CNPq, brazil) and Programa de Formação de Recursos Humanos (PRH-ANP, brazil). The authors would like to thank the VersusAI team for their invaluable support. Additionally, the authors acknowledge the use of generative AI language tools for light editing, such as spelling and grammar corrections, in the preparation of this manuscript. 

%% file: sections/8_references.tex
\bibliographystyle{./bibliography/IEEEtran}
\bibliography{./bibliography/IEEEabrv,main}